%% file: main.tex
\documentclass[manuscript]{acmart}
\usepackage{subcaption} 
\AtBeginDocument{%
  }

\setcopyright{acmlicensed}
\copyrightyear{2025}
\acmYear{2025}
\acmDOI{XXXXXXX.XXXXXXX}

\acmConference[LAK '25]{The 15th International Learning Analytics \& Knowledge Conference}{March 03--07,
  2025}{Dublin, Ireland}
\acmISBN{978-1-4503-XXXX-X/18/06}

\usepackage{geometry}
\usepackage{microtype}
\usepackage{ragged2e}
\usepackage{booktabs, multirow, xltabular}
\usepackage{bm}
\usepackage{subfiles}
\usepackage{soul}
\usepackage{changepage}

\begin{document}

\title[Applying IRT to Distinguish Between Human and Generative AI Responses to Multiple-Choice Assessments]{Applying Item Response Theory to Distinguish Between Human and Generative AI Responses to Multiple-Choice Assessments}

\author{Alona Strugatski}
\email{alona.faktor@weizmann.ac.il}
\orcid{0009-0001-3282-7067}
\affiliation{%
  \institution{Weizmann Institute of Science}
  \city{Rehovot}
  \country{Israel}
}

\author{Giora Alexandron}
\orcid{0000-0003-2676-6912}
\affiliation{%
  \institution{Weizmann Institute of Science}
  \city{Rehovot}
  \country{Israel}
  }
\email{giora.alexandron@weizmann.ac.il}

\renewcommand{\shortauthors}{Strugatski \& Alexandron}

\begin{abstract}
\textbf{ABSTRACT.}\\
Generative AI is transforming the educational landscape, raising significant concerns about cheating. Despite the widespread use of multiple-choice questions (\textbf{MCQs}) in assessments, the detection of AI cheating in MCQ-based tests has been almost unexplored, in contrast to the focus on detecting AI-cheating on text-rich student outputs. In this paper, we propose a method based on the application of Item Response Theory (\textbf{IRT}) to address this gap. Our approach operates on the assumption that artificial and human intelligence exhibit different response patterns, with AI cheating manifesting as deviations from the expected patterns of human responses. These deviations are modeled using Person-Fit Statistics (\textbf{PFS}). We demonstrate that this method effectively highlights the differences between human responses and those generated by premium versions of leading chatbots (ChatGPT, Claude, and Gemini), but that it is also sensitive to the amount of AI cheating in the data. Furthermore, we show that the chatbots differ in their reasoning profiles. Our work provides both a theoretical foundation and empirical evidence for the application of IRT to identify AI cheating in MCQ-based assessments.
\end{abstract}



\newcommand\blfootnote[1]{%
  \begingroup
  \renewcommand\thefootnote{}\footnote{\textcolor{red}{#1}}%
  \addtocounter{footnote}{-1}%
  \endgroup
}
\blfootnote{PRE-PRINT VERSION}
\blfootnote{Accepted to The 15th International Learning Analytics and Knowledge Conference (LAK'25).}



\maketitle

\section{Introduction}
Generative artificial intelligence (GenAI), and specifically conversational chatbots such as ChatGPT, are 
disrupting the educational landscape  \cite{yan2024practical}. 
One of the most controversial issues surrounding the use of GenAI in education is that it makes cheating alarmingly easy, leading many educators to fear it will result in widespread counterproductive and unethical learning behaviors \cite{khalil2023will, cotton2024chatting,The-AI-Cheating-Crisis}. Large-scale cheating poses significant risks to education. First, it undermines the validity of assessments, potentially leading to incorrect decisions about students (e.g., in college admissions) \cite{singhal_82}.  Second, as cheating erodes confidence in the reliability of assessments as accurate measures of learners' abilities, it can ultimately devalue academic credentials \cite{CAE17}. Third, learning-wise, cheating can negatively affect the learning process \cite{alexandron2020assessment}. 
And since cheating tends to spread—either by altering social norms or by causing students to fear falling behind -- its impact intensifies over time \cite{mccabe2001cheating}.

Unfortunately, there is already ample evidence on the use of GenAI for cheating in various educational contexts \cite{Prothero24}, and this was marked as a major risk to the integrity of educational assessment \cite{susnjak2024chatgpt}. Most of the attention was given the cheating potential of GenAI on text-rich tasks such open questions or essays \cite{khalil2023will}. To tackle this, various methods have been proposed, which are mainly based on text similarities, and are implemented in commercial AI writing detection technology such as iThenticate\footnote{https://www.ithenticate.com/}  and Turnitin\footnote{https://www.turnitin.com/}.

However, educational assessment -- particularly large-scale, standardized, and high-stakes exams -- relies heavily on closed-form question formats, which are amenable to automated grading (in this discussion, we focus specifically on MCQs, the most common format). ChatGPT and the like have been shown to be very successful in solving the MCQ parts of engineering and science undergraduate exams from elite institutions \cite{borges2024could}, as well as on various bar exams  \cite{katz2024gpt}. The risk that `cheating with GenAI' poses to assessment has led a recent survey to conclude that ``all summative assessments using MCQs should be conducted under secure conditions with restricted access to ChatGPT and similar tools'' \cite{newton2024chatgpt}. Nevertheless, research on detecting the use of GenAI for answering MCQs is scarce, and in fact we are familiar with only one research addressing this issue \cite{sorenson2024identifying}. 

At first glance, detecting that GenAI was used to produce a response that is simply a digit seems implausible, as it contains very little information. The crux may be to look for patterns that emerge in sequences of responses, thus providing additional information that arises from the interactions between items. Examined from a psychometrics perspective, due to the fundamental architectural differences between human and artificial intelligence, we can expect these `cognitions' to be impacted differently by various task dimensions \cite{pellert2023ai}. As a simple illustration, let us consider conversational chatbots such as ChatGPT (hereafter, we will use conversational agents as the exemplifying application of GenAI, with ChatGPT serving as a representative example). ChatGPT is still limited in its ability to solve questions that involve figures or diagrams \cite{borges2024could}. This means that including a visual representation may impact ChatGPT in different ways than it would impact typical human learners (in the Discussion we refer to the fact that humans as well may not be impacted in the same way by different problem dimensions). But there are also differences related to conceptual understanding and reasoning. For example, \cite{wang2024examining} reported that ChatGPT was less accurate when requested to solve under-specified physics problems or to make assumptions about the real world. Other studies that examined Chatbots' problem solving ability in chemistry \cite{watts2023comparing} and biology \cite{duong2024analysis} reported that the GenAI struggled in ways that are different than human learners. These studies provide empirical evidence that the anticipated differences are indeed evident in real-life assessment data. 
However, in order to develop tools for detecting cheating using GenAI based on these differences, which can be generalized across contexts, it is advisable to rely on rigorous theories of assessment and measurement of cognitive abilities. A key theory of assessment that seems to be appropriate is IRT, which is a psychometric framework used to model the relationship between individuals' responses to assessment items and their underlying traits or abilities \cite{DeAyalaIRT}. 
The theoretical assumption behind IRT is that a plausible response pattern arises primarily from the interaction between the test-taker's ability and the difficulty of the items \cite{DeAyalaIRT}.  Deviations from the expected pattern suggest that the responses may have been influenced by an alternative response process such as guessing \cite{felt2017using} or cheating \cite{karabatsos2003comparing,alexandron2019towards, alexandron2023general,tendeiro2012cusum,guo2010identifying}. As GenAI as well is an `alternative' response pattern, our hypothesis was that its responses would deviate from the expected response patterns. The question is whether it is possible to model these deviations using IRT in a way that highlights the differences between humans and artificial intelligence. Recently it was suggested that misfit in IRT modeling can indicate AI cheating \cite{sorenson2024identifying}, but the idea was tested using a narrow measure (outfit statistics) computed by a proprietary IRT package and tested solely on ChatGPT 3.5.

We propose to apply the general theoretical framework of PFS. Within IRT, PFS is a collection of statistical methods designed to assess the alignment between an individual's test performance and an expected response pattern \cite{meijer2001methodology} (see Subsection~\ref{SubSecPersonFit} for more details on PFS). Following this, the first goal of our research (as formulated through the first research question; see Subsection~\ref{subsec:RQs}) is to examine whether IRT and PFS can highlight differences between the response patterns of human learners and GenAI in MCQ-based assessments. We examined this in two very different assessment contexts: A high-school chemistry instrument administered to about one thousand high-school students as \textit{formative} assessment, and the responses of a few thousand examinees to a quantitative chapter from  a
\textit{summative}, high-stakes psychometric test (undergraduate entrance exam).  To obtain the GenAI responses, we asked the premium versions of three major conversational chatbots -- ChatGPT-4o, Gemini 1.5 pro, and Claude 3.5 sonnet -- to solve these instruments. Each chatbot was run 20 times to mimic the responses of 20 students who used the chat for cheating. We then computed the PFS for the human and GenAI responses (see Section~\ref{SecMethod} for methodological details). 

Foreshadowing the results, we show that using IRT and PFS (see Subsection ~\ref{SubSecPersonFit} for the exact PFS measures  we use) indeed reveals  statistically significant differences between human learners and conversational chatbots. But since different chatbots (e.g., ChatGPT, Gemini, etc.) may vary substantially in their response patterns due to architectural differences, the data used for their training, and various types of alignment mechanisms, it also raises the question of whether `chatbots' are actually an homogeneous group or not. Studying this question is the second goal of our research (formulated through the second research question). The findings  show that the three chatbots -- ChatGPT-4o, Gemini, and Claude -- differ significantly in their response patterns, as measured by PFS, indicating that, at least from an IRT perspective, \textit{these tools exhibit different reasoning processes}.

PFS measure aberrancy. As long as cheating is not the norm, it appears as an aberrant response pattern. However, if it becomes more prevalent and less `aberrant,' anomaly-based techniques become less effective at detecting it \cite{karabatsos2003comparing,alexandron2019towards, alexandron2023general}. This raises a practical question: under which conditions -- and more precisely, amount of cheating -- GenAI responses are still distinguishable? 
We study this through the third question that our research seeks to answer. We follow the sensitivity analysis scheme of  \cite{karabatsos2003comparing}, which examined 36 PFS for identifying cheating under increasing levels of pollution (namely, proportion of cheaters within the learner population).

The main contribution of this paper is the presentation of an IRT and PFS-based learning analytics approach for detecting the use of GenAI to cheat in MCQ-based educational assessments. The approach is grounded in well-established educational measurement theory, and its effectiveness in distinguishing between human learners and conversational chatbots, as well as between different types of chatbots, is demonstrated. Additionally, the study examines the limitations of these techniques, which depend on quantifying irregular cognitive patterns. However, as cheating becomes more widespread, these patterns shift from being aberrant to becoming the norm. More generally, the paper identifies educational measurement theory as a conceptual framework for distinguishing between humans and GenAI.

\section{Methodology}\label{SecMethod}

\subsection{Research Questions}\label{subsec:RQs}
The goals of of our research are formulated into the following research questions (RQs):\\
\textit{\textbf{RQ1}}: Do the response patterns of conversational chatbots differ from those of human learners, as measured by PFS?\\
\textit{\textbf{RQ2}}: Are there substantial differences between the response patterns of different conversational chatbots?\\
\textit{\textbf{RQ3}}: What is the impact of the level of pollution on the difference between the PFS measures of GenAI and human learners?

\subsection{Procedure and Data}\label{SubSecProcedure}\label{SubSecData}  
To study the capabilities of IRT in differentiating between the response patterns of human learners and GenAI in MCQ assessments, we use datasets that include the responses of human learners in two assessment contexts: science education and a standardized high-stakes exam. By introducing a small percentage of AI-generated responses (from ChatGPT, Claude, and Gemini, with 20 responses each), we assess the effectiveness of IRT-based PFS -- \(G\), \(G^*\), \(U3\), and \(ZU3\) -- to separate between the human and the GenAI responses, as well as between the different chatbots. Among the various PFS proposed in the literature (for a comprehensive list, see \cite{karabatsos2003comparing}), we have chosen these metrics as they have been shown to be the most effective in identifying cheating \cite{karabatsos2003comparing, alexandron2019towards, alexandron2023general}.

\subsubsection{Procedure} To answer RQ1 we used 2x3 datasets. 
The design included the two different instruments with students answers combined with 5\%  of responses from three different GenAI agents. For a given dataset, we calculated 4 PFS for each examinee: \(G\) \cite{guttman1944basis}, \(G^*\) \cite{van1977environmental}, \(U3\) and \(ZU3\) \cite{VanderFlier1980}. The statistics were calculated using the R package \textit{PerFit} \cite{PerFit}. 
Given the values, we used the Wilcoxon rank-sum test to compare the distribution of the statistics between the group of human agents and GenAI agents. 

To answer RQ3 we use similar 2x3 datasets as for RQ1, now using 3 level of pollution 5\%, 10\%  and 25\% following the framework of \cite{karabatsos2003comparing}. To create these datasets, we maintained all responses collected from the GenAI agents and combined them with the correct number of human responses to achieve the required level of pollution (more details in \ref{subsecRQ3_results}). Again, we used the Wilcoxon rank-sum test to compare the four measures between the two groups, humans and GenAI, at different pollution levels.

To answer RQ2: Are there substantial differences between response patterns of different chatbots on MCQ assessment, we compared the four PFS between the three chatbots using the Kruskal-Wallis test. We constructed two datasets of human answers (one per instrument), with  about 5\% of GenAI responses divided equally among the three chatbots.


\subsubsection{Instruments and student data}
We used two assessment instruments from very different contexts. The first was a chemistry diagnostic examination administered via a Moodle platform as a formative assessment activity in preparation for a high-school matriculation test. It included 22 MCQs, with 931 high-school student respondents available. The second instrument was a quantitative chapter from a psychometric exam taken by prospective students as a prerequisite for admission to higher education institutions. This instrument included 20 MCQs with over 4,800 respondents. Both instruments were multi-modal, containing for some of the questions images/figures and formulas.  

\subsubsection{Collection of Chat data}
The responses were generated using three different GenAI models: ChatGPT-4o, Gemini 1.5 pro, and Claude 3.5 sonnet. We uploaded a document for each instrument with a prompt that required only the final answers to the items, omitting any additional information or comments. The models typically provided an enumerated list of answers, which were then exported to a CSV format. This procedure was repeated 20 times for each model, simulating 20 `artificial students' per model. Each run was executed as a new session to prevent the models from leveraging multiple trials on the same instrument. Although the specific interaction with each model was prompted with the same text, the responses varied due to the temperature parameter being set to a non-zero value. All the chat models supported figures and visual modalities, which was important since our instruments contain some items that rely on multi-modal inputs. When the chat did not explicitly use one of the possible answers or skipped answering an item, we counted the response as incorrect.

\subsection{Person-fit statistics and their application to cheating detection }\label{SubSecPersonFit}
Miejer and Sijstma \cite{meijer2001methodology} presented in their review a large number of statistics invented for the purpose of identifying aberrantly responding examinees. Those PFS are may be classified to either parametric or non-parametric methods. The parametric models are based on IRT, and measure the distance between the test set, and the predicted responses derived from the IRT model with parameters fine-tuned on train data. A non parametric PFS are calculated directly from the data set on \(N\) examinees, and theirs scored responses to \(J\) items, without relaying on estimating parameters for IRT model \cite{karabatsos2003comparing}. Such non parametric PFS measures \(G\), \(G^*\) \cite{van1977environmental}, \(U3\) and \(ZU3\) \cite{VanderFlier1980} are used in this work. Let us have an \(\bm{X} = (X_{nj} \mid n = 1, \ldots, N, j = 1, \ldots, J)\), a set of examinee responses, where \(X_{nj} = 1\) denotes a correct response made by examinee \(n\) (human or GenAI agent) on item \(j\), and \(X_{nj} = 0\) denotes an incorrect response.
Calculating \(G\) and \(G^*\) depends on reasonableness of examinee \(n\), which defined as \(r_n=\sum_{j=1}^JX_{nj}\) and represents the sum of correct answers. The statistic G counts the number of item response pairs that deviate
from the ideal pattern. Such a pattern contains correct answers to the easiest \(r_n\) test items out of \(J\), and incorrect to the remaining, and  we refer it as ``Guttman pattern'' and denote this random vector as \(X_+\). The statistic \(G^*\) normalizes \(G\) to have the range [0,1]. 

Given ``Guttman pattern'' and ordered items by increasing value of difficulty, the ``reversed Guttman pattern'' is a vector with 1s on the last \(r_n\) positions, and 0s elsewhere \cite{meijer2001methodology}. \(U3\) Statistic is zero if the response vector has ``Guttman pattern'' and one for a response vector with the ``inverse response pattern.'' It is normally distributed with conditional expectation value and variance, conditioned on \(X_+\) for an examinee with reasonableness \(r_n\). \(ZU3\) transforms \(U3\) to have a unit-normal distribution (i.e., with mean = 0 and sd = 1). 

After computing the PFS for all examinees in the dataset, examinees with unusually high scores on any of these statistics are flagged as potential aberrant respondents.
For example, if an examinee's \( G^* \) or \( ZU3 \) is significantly higher than a pre-determined threshold (based on the normal distribution of these statistics), their response pattern is considered suspicious. The effectiveness of PFS improves with a greater number of items. Short tests may not provide enough data to reliably detect aberrant patterns. Typically, tests with at least 20 items show a reasonable variance, that allows anomaly detection \cite{meijer-review, VanderFlier1980}. The instruments used in this study contain at least 20 items. The explicit form of the four statistics used in this work can be found in \cite{karabatsos2003comparing}.

\section{Experiments and Results}\label{SecResults}
\subfile{Experiments}

\section{Discussion}\label{SecDiscussion}
Along with the great potential of GenAI, there is a growing fear that it will lead to large-scale cheating, devaluing educational assessment. Therefore, solutions that prevent or reduce GenAI cheating are very important. However, while MCQs are arguably the most common questioning format, there has been very little research on preventing GenAI cheating in MCQ-based assessments.
We propose building systems based on educational measurement theory, specifically IRT and PFS, to model assessment behaviors using metrics that are sensitive to the differences between human and artificial intelligence behavior. In two assessment contexts and on a few major conversational agents, we show that  PFS measures can be used to capture the different response patterns of humans and GenAI (RQ1). However we also show that chatbots differ in terms of their assessment behavior (RQ2), and that the effectiveness of PFS-based measures in distinguishing between humans and GenAI decreases as GenAI becomes more prevalent (RQ3). Thus, while applying IRT and PFS to separate between humans and GenAI is a promising approach, much more research in this direction is needed, especially given the heterogeneity of human learning \cite{salman24}, which implies that human learners as well are expected to differ significantly from one another. 

\noindent\textbf{\textit{Contribution}.} The main contribution of this paper is proposing a conceptual framework that applies well-established educational measurement theory -- IRT, and within it, PFS -- for distinguishing between GenAI and humans in MCQ-based educational assessments, demonstrating its effectiveness in two very different, authentic assessment contexts, and on three leading conversational agents. Additionally, the study explores the limitations
of such techniques.

\noindent\textbf{\textit{Limitations.}} 
While promising, PFS-based cheating detection methods have some key limitations. First, they are sensitive to the level of pollution, as demonstrated in RQ3. Second, they are applied retrospectively and less appropriate for real-time detection. Third, misfit that leads to high PFS measures can also arise from legitimate conditions such as learning disabilities.  Regarding research validity, the main limitation of this research relates to its external validity, as it was applied to a small number of instruments, and examined three specific GenAI tools. However, this research is ongoing and aims to pave the way for more larger-scale research on applications of IRT to the challenge of distinguishing between human and GenAI. 

\noindent\textbf{\textit{Future work.}}
The present paper centered on establishing the conceptual and empirical foundation for using IRT and PFS to identify GenAI-generated responses in MCQ assessments. In future work, we intend to build on this foundation to develop machine learning algorithms for detecting GenAI-assisted cheating.

\begin{acks}
This research was supported by the Israeli Ministry of Innovation, Science and Technology, Project number 1001707050.
The authors thank the National Institute for Testing and Evaluation for providing access to psychometric exam data.
\end{acks}

\bibliographystyle{ACM-Reference-Format}
\bibliography{MyBib}

\vspace{0.2cm} 
\appendix
\subfile{Appendix_A}

\end{document}

%% file: Experiments.tex
\subsection{Comparing the Response Patterns of Conversational Chatbots and  Human Learners (RQ1)}\label{subsec:RQ1_results}

As described above, we compared four PFS (\(G\), \(G^\ast\), \(U3\), and \(ZU3\)) on the 2x3 datasets including responses from human learners and 3 GenAI chatbots (ChatGPT, Gemini, and Claude) on 2 assessment instruments (chemistry and psychometric quantitative test). 
The PFS were computed for each combination (\textit{chatbot X instrument}) in separate, with a 5\% pollution level, meaning that 5\% of the `students' in each dataset were from the chatbot. 

\begin{figure}
    \centering
    \includegraphics[width=0.85\linewidth]{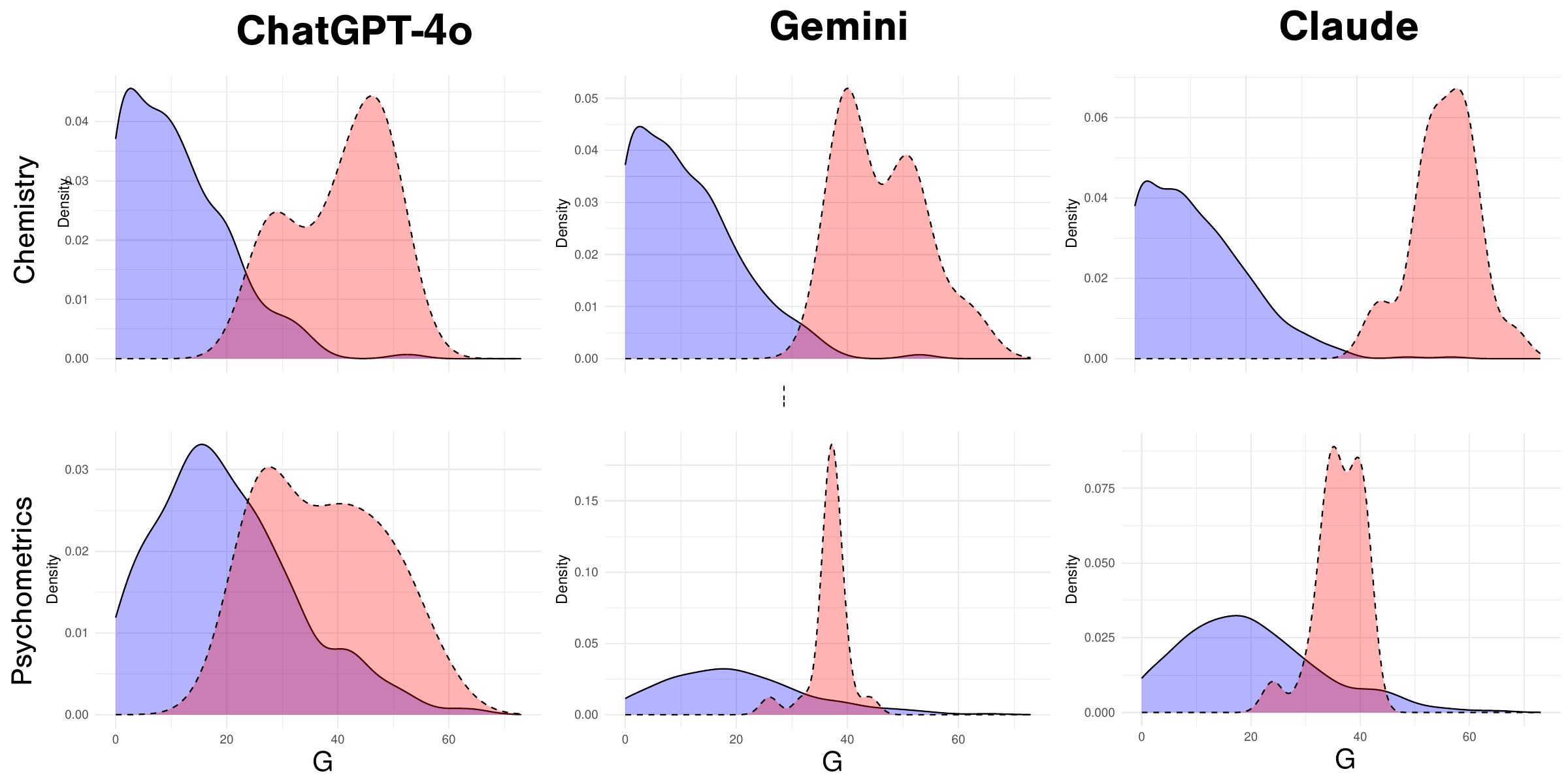}
    \captionsetup{justification=centering} 
    \caption{Density plots of \(G\) for human learners (blue) and conversational chatbots (red).}
    \label{fig:RQ_1}
\end{figure}

Table \ref{tab:RQ1_res} presents the results of the one-sided Wilcoxon signed-rank test for these comparisons. Notably, the tests show that the PFS of the chatbots were statistically significantly higher than those of the human respondents, with the highest p-value being less than 0.00001. This finding held consistently across all the 2x3x4 combinations of instruments, AI agents, and PFS measures, confirming our hypothesis. This is demonstrated in Figure~\ref{fig:RQ_1}, which shows the well separated density plots of the \(G\) PFS of human learners and the chatbots. As can be seen, the \(G\) measure effectively distinguishes between these two groups, showing its value in assessing the response source. Additional figures displaying the density distributions for \(G^\ast\), \(U3\), and \(ZU3\) can be found in appendix A.

Thus, with respect to RQ1, we conclude that the response patterns of conversational chatbots differ significantly from those of human learners, as measured by IRT and PFS.

\begin{table}[ht]
    \centering
    \begin{tabular}{||c c|c c c c |c c c c|c c c c ||}
    \hline
        & & \multicolumn{4}{c|}{\textbf{ChatGPT-4o}} & \multicolumn{4}{c|}{\textbf{Gemini}} & \multicolumn{4}{c||}{\textbf{Claude}} \\ 
        \textbf{}& {}&\textbf{\(G\)}< & \textbf{\(G^*\)}< & \textbf{\(U3\) <} & \textbf{\(ZU3\)<}& \textbf{\(G\)<} & \textbf{\(G^*\)<} & \textbf{\(U3\)<} & \textbf{\(ZU3\)<} &  \textbf{\(G\)<} & \textbf{\(G^*\)<} & \textbf{\(U3\)<} & \textbf{\(ZU3\)<}  \\ 
        \hline
       \multirow{2}{*}{p-value}&  Chem.  & \(10^{-12}\) & \(10^{-10}\) & \(10^{-10}\) & \(10^{-12}\) &\(10^{-13}\) & \(10^{-11}\)& \(10^{-11}\)& \(10^{-13}\)& \(10^{-13}\)& \(10^{-11}\) & \(10^{-11}\)& \(10^{-12}\)\\ 
        & Psych. & \(10^{-7}\) & \(10^{-5}\) & \(10^{-6}\) & \(10^{-6}\)  & \(10^{-8}\)& \(10^{-6}\)& \(10^{-6}\)& \(10^{-4}\)& \(10^{-8}\) & \(10^{-6}\)& \(10^{-7}\)&\(10^{-6}\) \\ 
        \hline
       
         \multirow{2}{*}{Z}& Chem.  & -7.336 & -6.805 & -6.713 & -7.147 & -7.57 & -7.014& -7.005& -7.327&-7.615& -7.127& -7.007&-7.476\\ 
        & Psych.  & -5.445 & -4.764& -5.234& -5.084&-5.927&-5.051&-5.076& -4.415&-5.778& -5.1057& -5.424& -5.287\\ 
        \hline
  
        \hline
    \end{tabular}
    \captionsetup{justification=centering} 
    \caption{Wilcoxon test results for different measures across GenAI models ChatGPT-4o, Gemini, and Claude.}
    \label{tab:RQ1_res}
\end{table}

\subsection{Analysis of Person-Fit Statistics Measures of Different GenAI Respondents (RQ2)}\label{subsecRQ2_results}

To address RQ2, we investigated whether there are substantial differences between the PFS values (\(G\), \(G^\ast\), \(U3\), and \(ZU3\)) of the responses of the ChatGPT, Claude, and Gemini `students' on the 2 instruments. 
The datasets were constructed in the following manner: for the chemistry instrument, the dataset contained 976 responses, with 931 responses from human students and 45 responses from the three GenAI agents (15 responses from each agent). For the psychometric instrument, the dataset contained  1040 responses, with 980 responses from human students and 60 responses from the three GenAI agents (20 responses from each agent). This set-up aimed to maintain the experimental rationale of computing the PFS for 5\% pollution level.  We then compared the PFS of the three chatbots on each measure and each dataset using Kruskal-Wallis test. 

\begin{figure}
    \centering
    \includegraphics[width=1\linewidth]{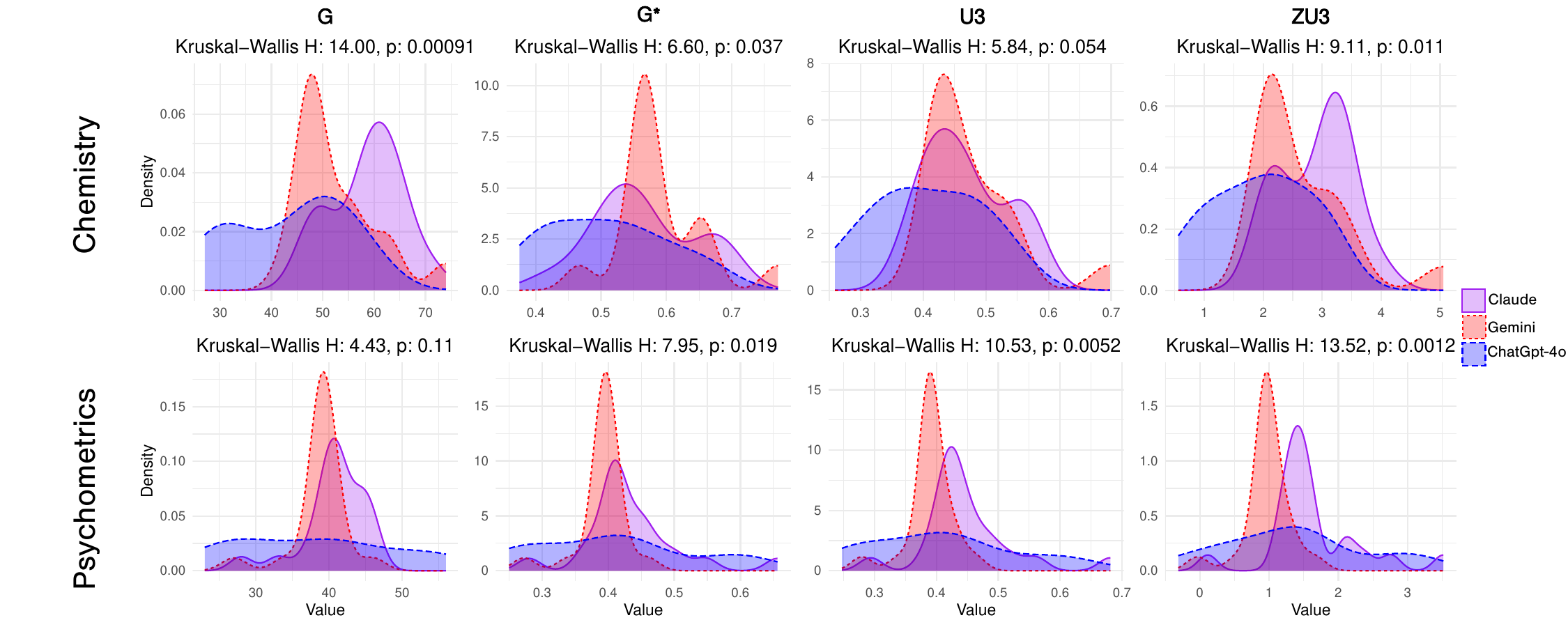}
    \caption{Results and density plots of PFS for conversational chatbots.}
    \label{fig:Measures_RQ2}
\end{figure}

\textbf{\textit{Results on the chemistry instrument:}} The Kruskal-Wallis test indicated significant differences in the PFS values among the three GenAI agents as follows: for \(G\) (H = 14.00, p < 0.001), for \(G^\ast\) (H = 6.60, p < 0.05) and \(ZU3\) (H = 9.11, p <0.05) . However, for \(U3\) (H = 5.84, p = 0.054), the differences were not statistically significant. Figure \ref{fig:Measures_RQ2} illustrates the density plots of the measures for different chats, providing a visual representation of the response pattern distributions across the GenAI agents.

To further explore these differences, we conducted a post-hoc analysis using Dunn test with Bonferroni correction. ChatGPT showed significant differences on \(G\) and \(ZU3\) when compared to both Gemini (\(G\): p = 0.030, \(ZU3\): p = 0.0190) and Claude (\(G\): p = 0.0004, \(ZU3\): p = 0.0276) .
For \(U3\),  where the Kruskal-Wallis result was borderline (p = 0.054), the Dunn test indicated that still ChatGPT was distinct from Gemini (p = 0.0254). Additionally, Gemini was distinct for \(G^\ast\), with the pairwise comparison with Claude and ChatGPT yielding p-values of 0.0015 and 0.0008, respectively. 

\textbf{\textit{Results on the psychometric instrument:}} The Kruskal-Wallis test similarly revealed significant differences among the three GenAI agents for \(G^\ast\) (H = 7.95, p < 0.05), \(U3\) (H = 10.53, p < 0.01) and \(ZU3\) (H = 13.52, p < 0.01). However, the differences for \(G\) (H = 4.43, p = 0.11) were not statistically significant.

The post-hoc analysis revealed distinct groupings for Claude and Gemini across all three measures (\(G^\ast\), p = 0.0149; \(U3\), p = 0.0039; and \(ZU3\), p = 0.0007). However, for the psychometric instrument, ChatGPT exhibited characteristics that placed it into both groups, being distinct from neither Claude nor Gemini. 

Summarizing the results of RQ2, we conclude that there are significant differences between the response patterns generated by different GenAI agents. However, these differences are not always captured by all the measures (\(G\), \(G^\ast\), \(U3\), and \(ZU3\)) and across instruments. Specifically, ChatGPT-4o stood as having the most distinct response pattern on the chemistry instrument, and Gemini as having the most distinct response pattern on the psychometric  instrument. 

\subsection{The Impact of Pollution Level on the Difference Between the PFS of GenAI and Human Learners (RQ3)}\label{subsecRQ3_results}
\begin{figure}
    \centering
   \includegraphics[width=1\linewidth]{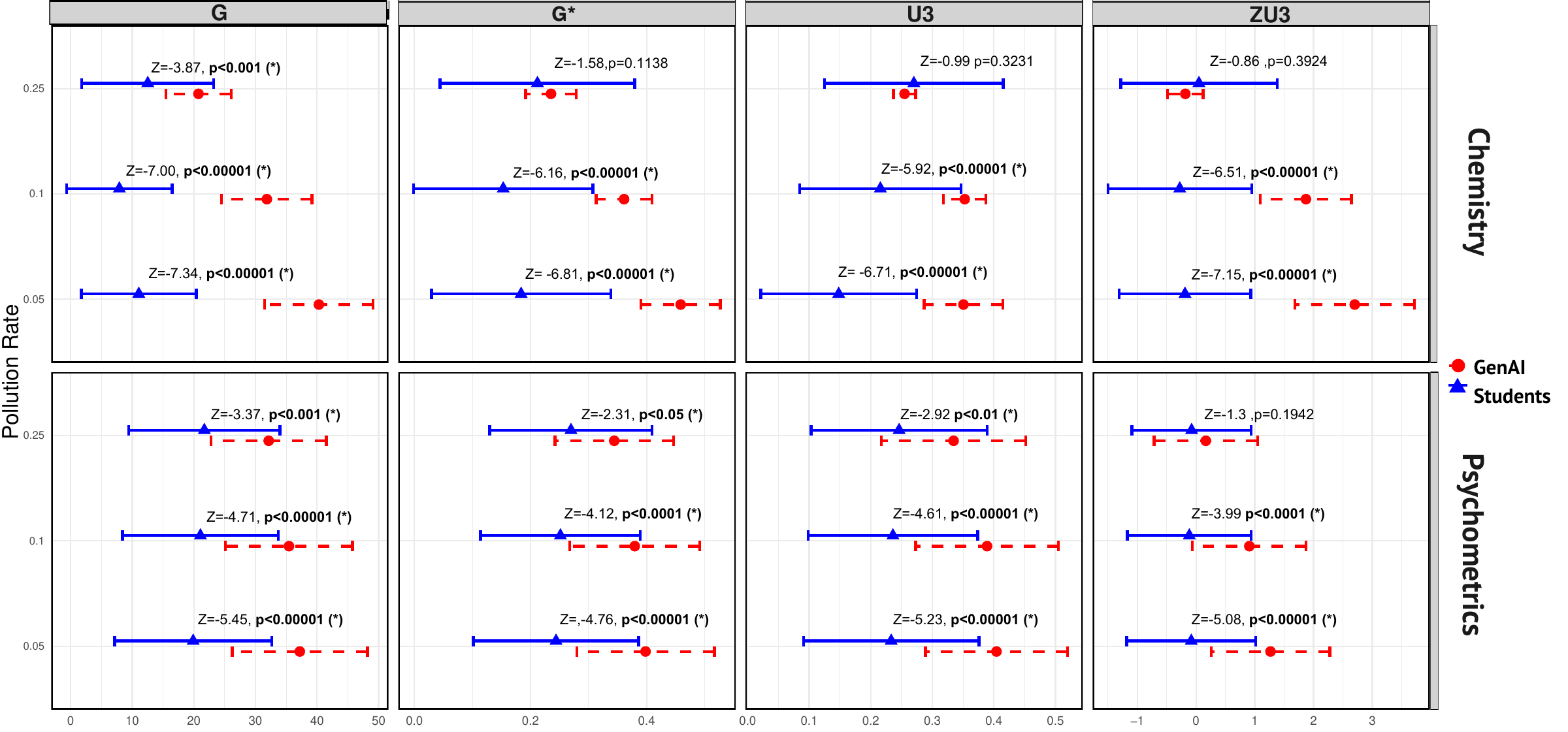}
    \caption{PFS mean with $\pm$1-STD range for ChatGPT with varying pollution rates for the chemistry and psychometric instruments.}
   \label{fig:RQ3}
\end{figure}

To address RQ3, we used 2x3 datasets as were used for RQ1, but included three levels of pollution: 5\%, 10\%, and 25\%, following the framework of \cite{karabatsos2003comparing}, resulting in 2x3\textbf{x3} datasets. To create these datasets, we maintained the 20 responses collected from each GenAI agent and combined them with the appropriate number of human responses to achieve the required level of pollution (380 learners' responses for 5\%, 180 for 10\%, and 60 for 25\%). Similar to RQ1, we used the Wilcoxon test to compare the four PFS measures between humans and GenAI at different pollution levels. 
The results are exemplified in Figure~\ref{fig:RQ3}, which presents the mean values with $\pm$1-STD of \(G\), \(G^\ast\), \(U3\), and \(ZU3\) for ChatGPT-4o on the chemistry and psychometric instrument under the three pollution conditions. As can be seen, the difference between these PFS values for the chatbot `students' and the human ones diminishes as the level of pollution increases. 

At pollution levels of 5\% and 10\%, the Wilcoxon results revealed a significant difference for ChatGPT, with p-value<0.0001 on all measures and on both instruments (the 5\% pollution is the same dataset as for RQ1). The results for  Claude are similar (not presented in the figure). For Gemini, while significant difference was evident for three out of the four PFS measures, \(ZU3\) showed an insignificant result at the 10\% pollution level (p-value=0.3655) indicating a potential sensitivity of \(ZU3\) to the level of pollution.

At a pollution level of 25\%, the differences between human learners and AI-generated responses for the \(ZU3\) measure become negligible across all chat models and instruments, with a \textit{minimal} p-value of 0.1942 for ChatGPT-4o in the psychometric test (see Figure \ref{fig:RQ3}). Similarly, for Claude and Gemini (not shown in the figure) across both instruments, as well as for ChatGPT using the chemistry instrument, measures \(G^*\) and \(U3\) also show no significant differences at this pollution level, with minimal p-value= 0.1138. 

Also evident in Figure~\ref{fig:RQ3} is that for ChatGPT, on \(G\), \(G^*\), and \(U3\) there were still statistically significant differences on the psychometric test with 25\% pollution. On the chemistry instrument, only on \(G\) there were statistically significant differences on this level for ChatGPT and for Claude (p-value < 0.0001). 

Overall, we conclude that as the level of pollution increases, the differences between the human and GenAI responses, as measured using PFS, diminishes. On the 25\% pollution level most of the differences are not statistically significant anymore, with \(G\) being the \textit{less} sensitive, and \(ZU3\) being the \textit{most} sensitive to the level of pollution. 

%% file: Appendix_A.tex
\textbf{Appendix A: Additional Figures for RQ1}\label{appendix_a}
\begin{figure}[ht!]
    \centering
    \begin{subfigure}{0.29\linewidth}
        \centering
        \includegraphics[width=\linewidth]{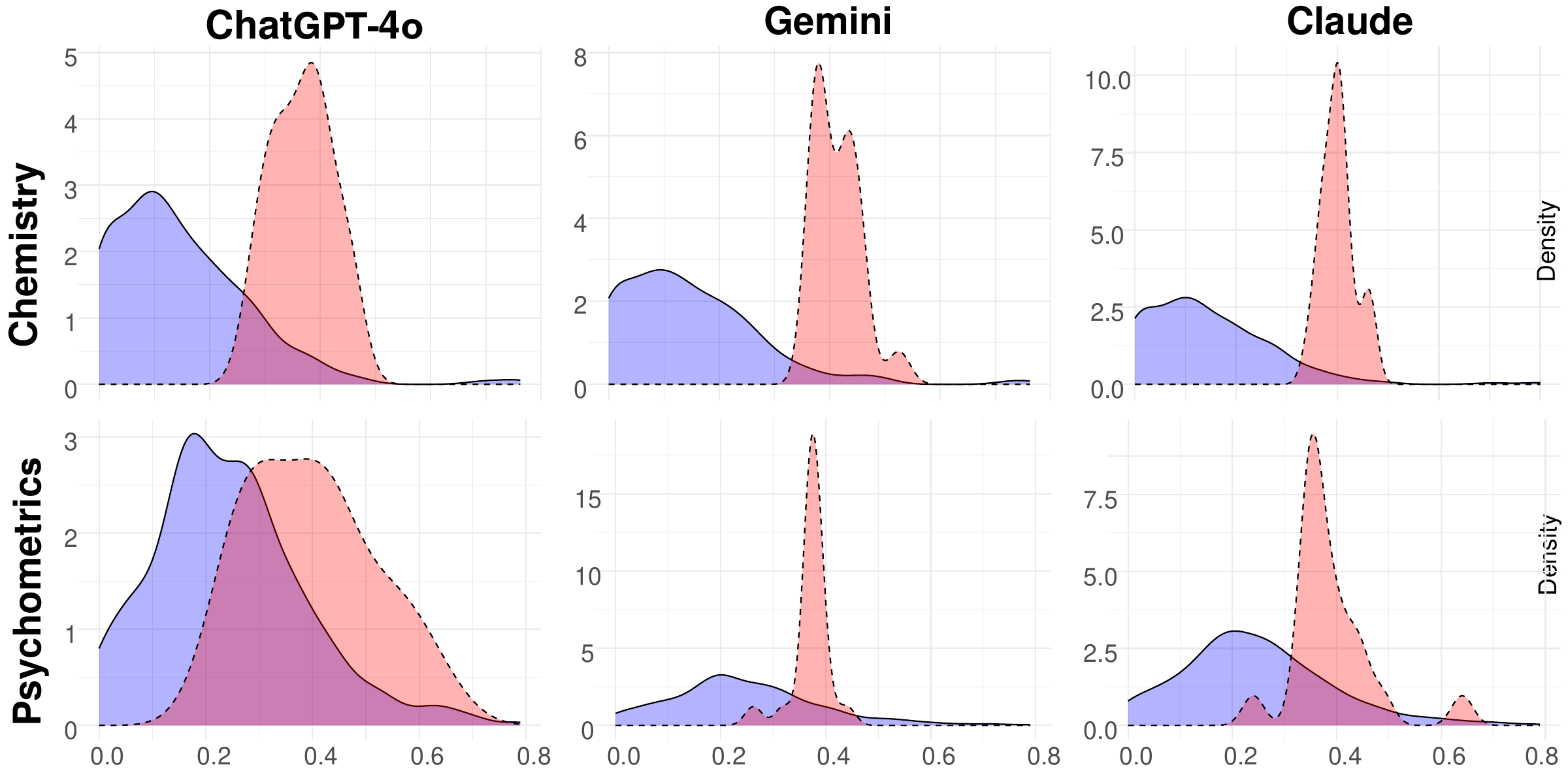} 
        \captionsetup{labelformat=empty} 
        \caption{\(G^*\)}
        \label{fig:sub1}
    \end{subfigure}
    \hfill
    \vrule{}
    \hfill
    \begin{subfigure}{0.29\linewidth}
        \centering
        \includegraphics[width=\linewidth]{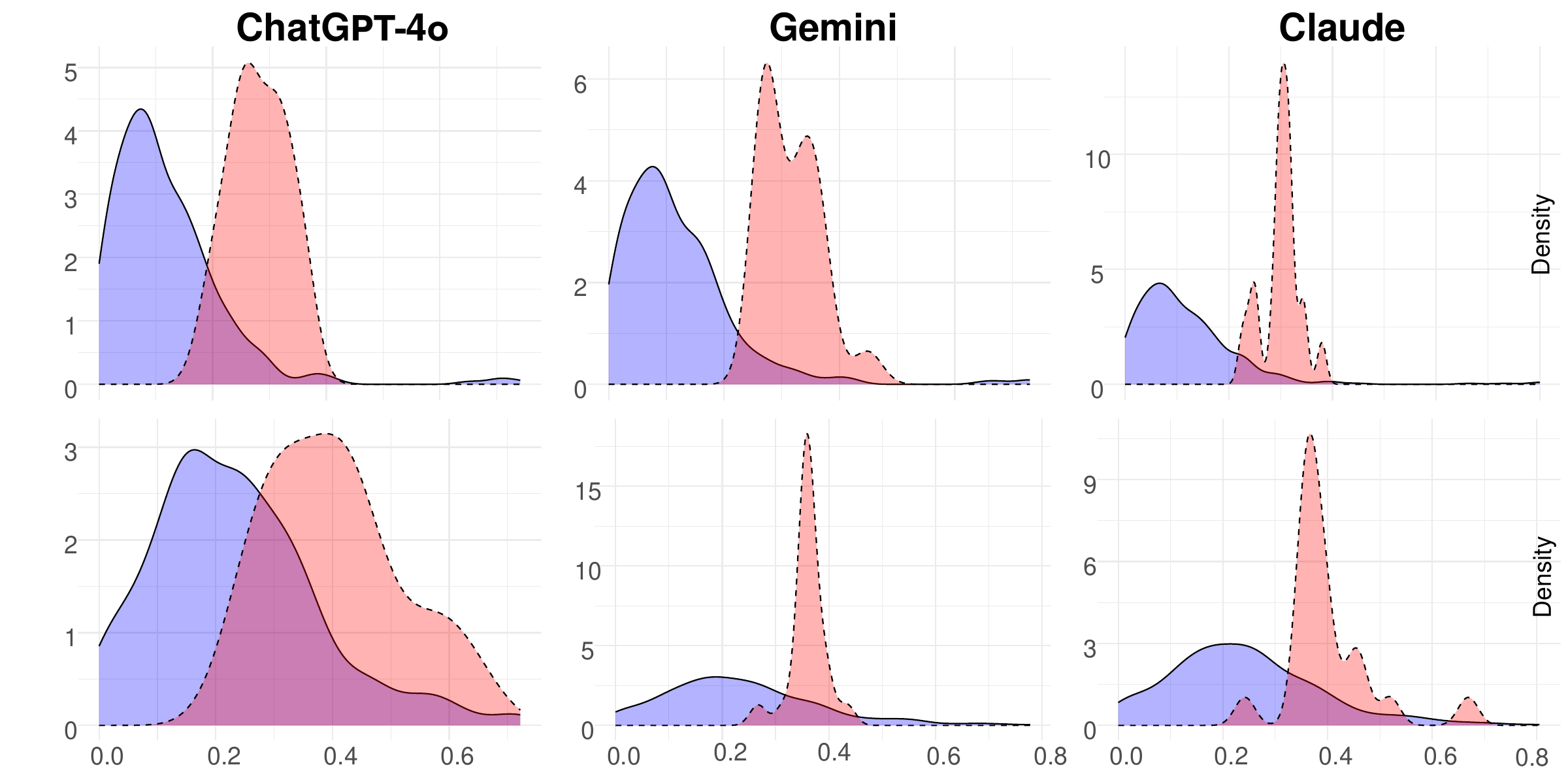} 
        \captionsetup{labelformat=empty} 
        \caption{\(U3\)}
        \label{fig:sub2}
    \end{subfigure}
    \hfill
    \vrule{}
    \hfill
    \begin{subfigure}{0.29\linewidth}
        \centering
        \includegraphics[width=\linewidth]{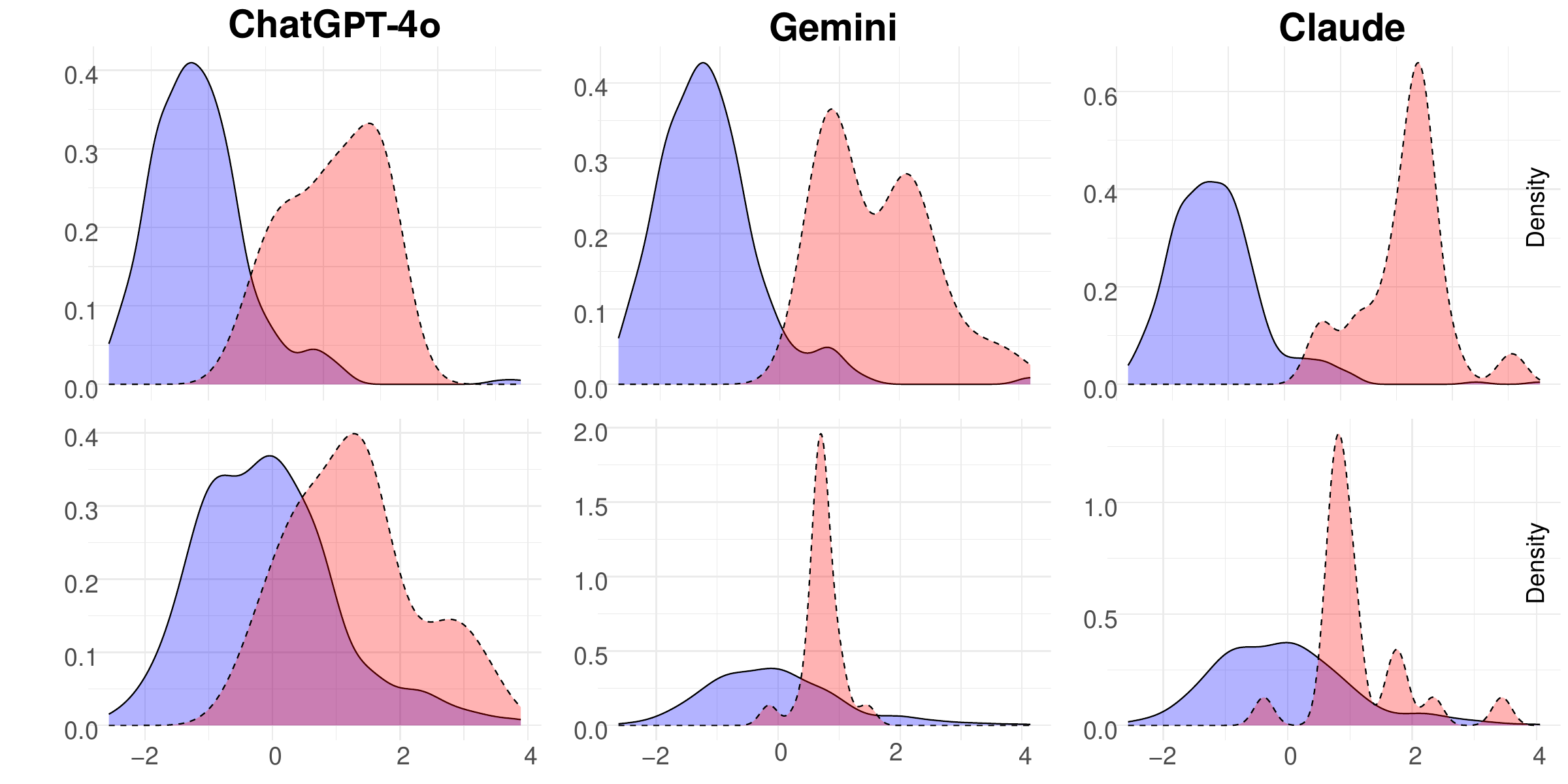} 
        \captionsetup{labelformat=empty} 
        \caption{\(ZU3\)}
        \label{fig:sub3}
    \end{subfigure}
    
    \caption{Density plots PFS for human learners (blue) and conversational chatbots (red). All calculated using 5\% of conversational chatbot responses. These figures show the difference between the human learners and the different conversational chatbots, for the two instruments used in this study - chemistry (formative) and psychometrics (summative).} 
    \label{fig:example}
\end{figure}